\pdfoutput=1

\documentclass[11pt]{article}

\usepackage{emnlp2021}

\usepackage{times}
\usepackage{latexsym}
\usepackage{amsfonts}
\usepackage{amsmath}
\usepackage{multirow}
\usepackage{graphicx}
\usepackage{booktabs}
\usepackage{tabularx}
\usepackage{bm}
\usepackage{floatrow}
\usepackage{multirow}
\usepackage{makecell}

\usepackage{adjustbox}
\usepackage{lipsum}  
\usepackage{tikz}
\usetikzlibrary{positioning}
\usetikzlibrary{backgrounds}
\usepackage{tikzscale}
\usepackage{pgfplots}
\pgfplotsset{width=10cm,compat=1.9}
\usepgfplotslibrary{groupplots}
\usetikzlibrary{pgfplots.statistics}
\usepackage{caption}
\usepackage{subcaption}

\usepackage[T1]{fontenc}

\usepackage[utf8]{inputenc}

\usepackage{microtype}

\begin{document}

\title{Perceived and Intended Sarcasm Detection with Graph Attention Networks}
  
  \author{Joan Plepi  \and Lucie Flek \\
 Conversational AI and Social Analytics (CAISA) Lab\\
Department of Mathematics and Computer Science, University of Marburg\\
 {\tt http://caisa-lab.github.io}\\}
  

\maketitle
\begin{abstract}
Existing sarcasm detection systems focus on exploiting linguistic markers, context, or user-level priors. 
However, social studies suggest that the relationship between the author and the audience can be equally relevant for the sarcasm usage and interpretation.
In this work, we propose a framework jointly leveraging (1) a user context from their historical tweets together with (2) the social information from a user's conversational neighborhood in an interaction graph, to contextualize the interpretation of the post. 
We use graph attention networks (GAT) over users and tweets in a conversation thread, combined with dense user history representations. 
Apart from achieving state-of-the-art results on the recently published dataset of 19k Twitter users with 30K labeled tweets, adding 10M unlabeled tweets as context, our results indicate that the model contributes to interpreting the sarcastic intentions of an author more than to predicting the sarcasm perception by others.
\end{abstract}

\section{Introduction}
Sarcasm is a form of non-literal language, in which the intended meaning of the utterance differs from the literal meaning, fulfilling a social function in a discourse \cite{doi:10.1080/01638539509544922,riloff-etal-2013-sarcasm}. Sarcasm detection poses a challenge for numerous NLP tasks, such as sentiment or stance prediction \cite{maynard-greenwood-2014-cares}.

Early sarcasm detection systems are based on lexical and syntactic cues \cite{10.1145/1651461.1651471,davidov-etal-2010-semi,Tsur_Davidov_Rappoport_2010,gonzalez-ibanez-etal-2011-identifying,10.1007/s10579-012-9196-x,ghosh-etal-2015-sarcastic}.
However, sarcasm interpretation requires context, even for humans \cite{wallace-etal-2014-humans}. More recent works hence incorporate discourse information such as contrast \cite{riloff-etal-2013-sarcasm,khattri-etal-2015-sentiment,joshi-etal-2015-harnessing,10.1145/2684822.2685316,tay-etal-2018-reasoning}, and contextualize the post by using features from user history \cite{bamman2015contextualized,amir-etal-2016-modelling,oprea-magdy-2019-exploring,hazarika-etal-2018-cascade}.
The relationship between an author and the audience has been given comparably less attention, despite its relevance for the sarcasm interpretation \cite{doi:10.1080/08824090109384781,doi:10.1080/10926488.2000.9678862,doi:10.1177/0261927X07309512,doi:10.1177/1461444810365313,bamman2015contextualized}. In this work, we propose a graph neural network framework jointly leveraging a user context from their historical tweets together with the social information from a user's neighborhood modeled by heterogeneous graph structures.


\textbf{The key contributions of this paper are:}

(1) We present the first graph attention-based model to identify sarcasm on social media by explicitly modeling users' social and historical context jointly, capturing complex relations between a sarcastic tweet and its conversational context. 

(2) We demonstrate that exploiting these relationships increases performance in the sarcasm detection task, reaching state-of-the-art results on the recent SPIRS dataset \cite{shmueli-etal-2020-reactive}, which we expand with user history. We examine the impact of different parts of the context, captured by attention weights, in modeling sarcastic utterances. 

(3) We find that even with user-based models, detecting sarcastic intentions of the author is easier than identifying the sarcasm perception by others.

\section{Related Work}



\paragraph{Leveraging user history}
Several previous works contextualize a sarcastic post by using features from user history - employing past tweets to identify a user's behavioral traits \citep{10.1145/2684822.2685316}, encoding user sentiment priors over different entities \citep{khattri-etal-2015-sentiment}, or manually crafting user interaction features \citep{bamman2015contextualized}. \citet{amir-etal-2016-modelling} introduce the user2vec model, applying paragraph2vec \citep{10.5555/3044805.3045025} over user history. \citet{hazarika-etal-2018-cascade} propose an alternative user embedding approach, encoding style and personality features.  

\paragraph{Leveraging user network}
An emerging line of research makes use of social interactions to encode information about the user induced by neural architecture \citep{10.1145/2939672.2939754, Qiu2018}. Network information improves performance on detecting cyberbullying \citep{10.1145/3292522.3326034}, abusive language use \citep{qian-etal-2018-leveraging}, suicide ideation \citep{mishra-etal-2019-snap} or fake news \cite{chandra2020graph}. To the best of our knowledge, graph network based approaches have not been used in the sarcasm detection task so far.

\paragraph{Perceived and intended sarcasm}
Perceiving sarcasm in text is not trivial even for humans, not only due to the lack of acoustic markers \citep{BANZIGER2005252, doi:10.1080/10926488.2011.583197} but also due to the sociocultural diversity \citep{doi:10.1080/08824090109384781, doi:10.1177/0261927X07309512} where in many cases the audience may misinterpret a sarcastic statement as sincere. 
This has been only recently reflected in sarcasm detection models \cite{hazarika-etal-2018-cascade,shmueli-etal-2020-reactive}.

\section{Proposed Approach}

\subsection{Tweet Embeddings}
We denote the current tweet to be assessed $t_i \in T = \{t_1, t_2, \dots , t_N \}$, where $N$ is the total number of tweets. 
We utilize SentenceBERT embeddings \cite{reimers-gurevych-2019-sentence} to encode the tweets. Formally,
    $\mathbf{t_i}^{\prime} = SentenceBERT(t_i)$
where $\mathbf{t_i}^{\prime} \in \mathbb{R}^{768}$, and SentenceBERT computes the mean of all tokens' representation. We forward this representation into a linear layer to transform in dimension $d$,  $ \mathbf{\widetilde{t_i}}  \in \mathbb{R}^d $.

\subsection{User Embeddings (Historical Context)}
Let $u^{t_i} \in U = \{u^{t_1}, u^{t_2}, \dots, u^{t_M}\}$ be the author of tweet $t_i$, from now on we keep only the index $i$ for brevity. Each user $u^i$ is associated with a set of historical tweets $\mathcal{H}^{i} = \{ (H_{1}^{i}, \tau_{1}^{i}), \dots , (H_{m}^{i}, \tau_{m}^{i})\}$, where $H_{j}^{i}$  is a historic tweet posted at a time $\tau_{j}^{i}$ by the user $u^{i}$. 
We adopt user2vec \cite{amir-etal-2016-modelling} to compute the initial user representation $\widetilde{\mathbf{u}_i} \in \mathbb{R}^d$ of user $u^{i}$ based on their corresponding historical tweets $\mathcal{H}^{i}$, optimizing the conditional probability of texts given the author.


\begin{figure}[!t]
\centering
\captionsetup{type=figure}
\includegraphics[width=\textwidth]{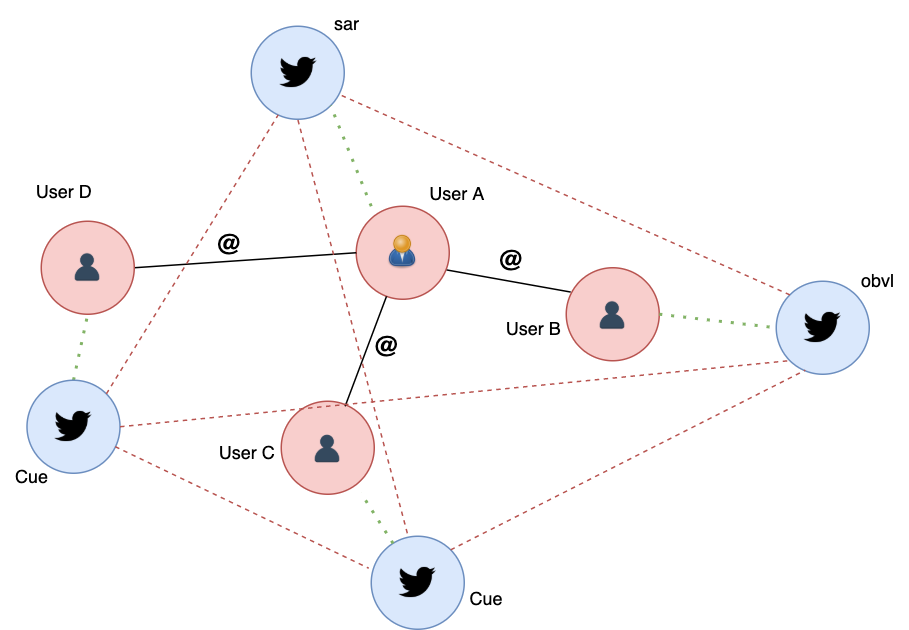}
\caption{An example of a heterogeneous user and tweet social graph extracted from one conversation.}
\label{fig:subgraph}
\end{figure}

\subsection{Social Graph (Network Context)}
\label{sec:socialgraph}
Apart from the importance of surrounding context to understand sarcasm \cite{wallace-etal-2014-humans},
certain understanding is needed between the audience and the author \cite{doi:10.1080/10926488.2000.9678862,doi:10.1177/0261927X07309512}.  Our goal is to model relations between users and their past tweets, interactions between users, and relations between tweets in one conversation. We model these relationships as a graph $ \mathcal{G} = (V, E)$, where $V = \{U \cup T\}$ contains two types of nodes - Users and Tweets (Figure \ref{fig:subgraph}). 
We use three edge types $E = \{ \mathrm{e}^U \cup \mathrm{e}^T \cup \mathrm{e}^C \}$, where $\mathrm{e}^U$ represents the social interaction between users. This involves quotes, mentions, or replies in the user history. $\mathrm{e}^T$ denotes the edges between tweets that are involved in one discussion thread, with all tweets connected with each other, and $\mathrm{e}^C$ is the relation between a tweet and its author. 

\paragraph{Representation Learning: } 
We use Graph Attention Networks (GATs, \cite{velickovic2018graph}) to exploit the neighborhood of each node to compute the final representations.\footnote{We ran early experiments with Graph Convolutional Networks as well, obtaining inferior and less interpretable results.} GAT uses a self-attention mechanism \cite{bahdanau2015neural, 10.5555/3295222.3295349} to assign an importance score to the connections that contribute more to the detection of sarcastic or non-sarcastic tweets. We initialize the user and the tweet nodes of the GAT with their corresponding embeddings
$\widetilde{\mathbf{u}_i}$ and $\widetilde{\mathbf{t}_i}$.
The initial node representation of each node $v \in V$ is linearly transformed by a weight matrix $ \mathbf{W} \in \mathbb{R}^{d^\prime \times d}$ into a vector $\mathbf{h}_v \in \mathbb{R}^{d^\prime}$. Following, the attention weights $e_{vn}$ of each node $v$ are computed as:
\begin{equation}
    e_{vn} = att(\mathbf{h}_v \| \mathbf{h}_n)
\end{equation}
\noindent where $n \in \mathcal{N}(v)$ is a node in the neighborhood of $v$ and $att$ is the attention mechanism function which is a single-layer feedforward neural network, parameterized by a weight vector $ \vec{\mathbf{a}} \in \mathbb{R}^{2d^\prime}$ with a LeakyReLU nonlinearity. 
\\

\noindent The final node representation $\mathbf{h}_{v}^{\prime} \in \mathbb{R}^{K \cdot d^{\prime}}$ is computed as: 
\begin{equation}
    \mathbf{h}_{v}^{\prime}= \sigma \left(\frac{1}{K} \sum_{k=1}^{K} \sum_{n \in \mathcal{N}(v)} \alpha_{vn}^{k}  \mathbf{W}^k   \mathbf{h}_{n} \right)
\end{equation}
\noindent where $K$ is the number of attention heads, $\sigma$ is the ReLU nonlinear function, $ \mathbf{W}^k \in \mathbb{R}^{d^\prime \times d^\prime}$ a weight matrix and $\alpha_{vn}^k = softmax(e_{vn}^k)$ the normalized attention weights from the $k$-th attention mechanism $att^k$.

\subsection{Classification model}
The user and tweet representations learned by GAT layer are concatenated and forwarded through a two-layer feed-forward network parameterized by weight matrices $\mathbf{W}_{1}^{c} \in  \mathbb{R}^{d_1 \times 2 d^\prime}$ and $\mathbf{W}_{2}^{c} \in  \mathbb{R}^{o \times d_1}$, where $d_1$ is the dimension of projected embeddings, and $o$ is equal to the number of classes. The final prediction of the model is given by: 

\begin{equation}
    \hat{y} = softmax \left(\mathbf{W}_{2}^{c} \left(\sigma \left(\mathbf{W}_{1}^{c} [h_t || h_u] \right) \right) \right)
\end{equation}

\begin{figure}[!t]
\centering
\captionsetup{type=figure}
\includegraphics[width=\textwidth]{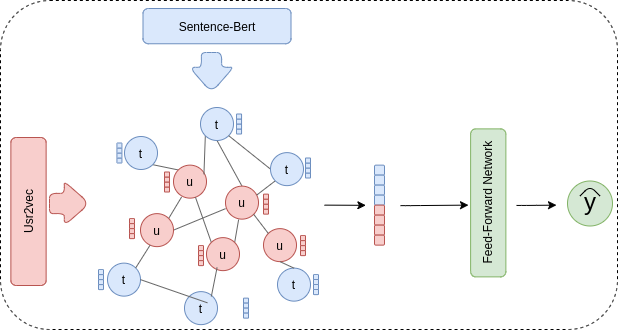}
\caption{The social graph is initialized with user and tweet embeddings (user2vec and sentence-{BERT}), and tuned by GAT to take into account relationships between them. The output representations are then fed into the classification layer.}
\label{fig:gat}
\end{figure}

\section{Experimental Setup}
\subsection{Dataset}
For our experiments, we use a recently published SPIRS sarcasm dataset \cite{shmueli-etal-2020-reactive}. It utilizes \textit{cue tweets}, conversation replies which point out the sarcastic nature of a previous post. 
In addition, the dataset also provides \textit{oblivious tweets}, questioning the sarcastic nature of a given example, and \textit{elicit tweets}, being the original start of the conversation. Non-sarcastic posts were collected randomly in equal numbers. The labeled dataset contains in total 15,000 sarcastic tweets (10,000 self-reported and 5000 perceived cues), 15,000 non-sarcastic, 10,000 oblivious and 9156 elicit tweets. 

\paragraph{User context} We extend SPIRS with over 10 million past tweets of the authors in the dataset in order to compute the user embeddings. 
\paragraph{Social network} Our graph consists of the three types of connections described in Sec.\ref{sec:socialgraph}.
To avoid the bias coming from cue tweets, we exclude these from our graph. Our final social network consists of 108K nodes with 0.00002 density and 32\% homophily, defined as the percentage of connections between authors of tweets with the same label.  

\subsection{Comparison Baselines}
The baselines introduced by \cite{shmueli-etal-2020-reactive} are a Convolutional Neural Network, a Bidirectional LSTM, and a fine-tuned pre-trained BERT model. We compare our model with BERT, which performs the best of these. We add two baselines which incorporate user information. First, we extend BERT by simply concatenating the tweet embeddings with their respective user2vec author representation (`BERT + user2vec'). As a second baseline (`BERT + user-only GAT'), we build a social graph with only user nodes and their interactions (quotes, mentions, or replies) $\mathrm{e}^U$ as edges, and apply the GAT initialized with user2vec embeddings. The implementation of the models and the results are made publicly available, to facilitate reproducibility and reuse\footnote{\url{https://github.com/caisa-lab/sarcasm_detection}}.  

\section{Results and Analysis}

Our proposed GAT base model significantly outperforms all the baselines (Table \ref{tab:detection_results}) despite having fewer trainable parameters (500K) than the BERT model (110M). 
First, by simply concatenating the user2vec embeddings to BERT, we obtain 3.4\% f1 score improvement on the BERT model, indicating the importance of user context in sarcasm detection. Moreover, we introduce the GAT module in the model. We first experiment with only tweet to tweet connections in the graph based on the conversations on Twitter and trained on top of the fine-tuned BERT. In this case, the GAT layer only bring 0.2\% improvement due to the sparse and disconnected nature of the constructed graph.  In addition, we replace user2vec with GAT embeddings tuned on user-only social graph, and we achieve 6.1\% improvement on BERT and 3\% over `BERT + user2vec', presumably thanks to exploiting the homophily relations between users. Finally, applying GAT on the full heterogeneous user and tweet graph (as per Figure \ref{fig:gat}) provides a large performance boost thanks to incorporating the conversational thread context between tweets.

\begin{table}[tpb]
\small
\centering
\renewcommand{\arraystretch}{1.5}
\begin{tabular}{lccc}
\toprule
\multicolumn{4}{c}{\textbf{Sarcasm Detection}  }               \\
\midrule
\textbf{Model}                 & \textbf{P} & \textbf{R} & \textbf{F1} \\ \midrule
BERT                          & 70.1\%     & 69.7\%     & 69.9\%      \\
                                    BERT + user2vec                & 73.6\%     & 73.2\%     & 73.4\%      \\
                                    BERT + tweet-tweet GAT                         & 70.4\%     & 69.9\%     & 70.1\%      \\
                                    BERT + user-only GAT                    & 74.2\%     & 78.1\%     & 76.1\%      \\
                                    \textbf{User+tweet GAT} (no cues)  & \textbf{84.7}\%     & \textbf{83.7}\%     & \textbf{84.2}\%      \\
                                    \hline

                                    User+tweet GAT, no elicit           & 83.2\%     & 80.8\%     & 82.0\%      \\
                                    User+tweet GAT, no oblivious        & 82.4\%     & 80.4\%     & 81.4\%      \\ 
                              User+tweet GAT + cue tweets              & 94.7\%     & 94.3\%     & 94.5\%      \\
                                    \bottomrule
                                    
\end{tabular}
\caption{Mean overall precision (P), recall (R), and F1 score (F1) of each model over 10 runs with varying seeds, detecting sarcasm on the SPIRS dataset. 
}
\label{tab:detection_results}
\end{table}

\begin{figure}[!t]
\centering
\captionsetup{type=figure}
\includegraphics[width=0.9\textwidth]{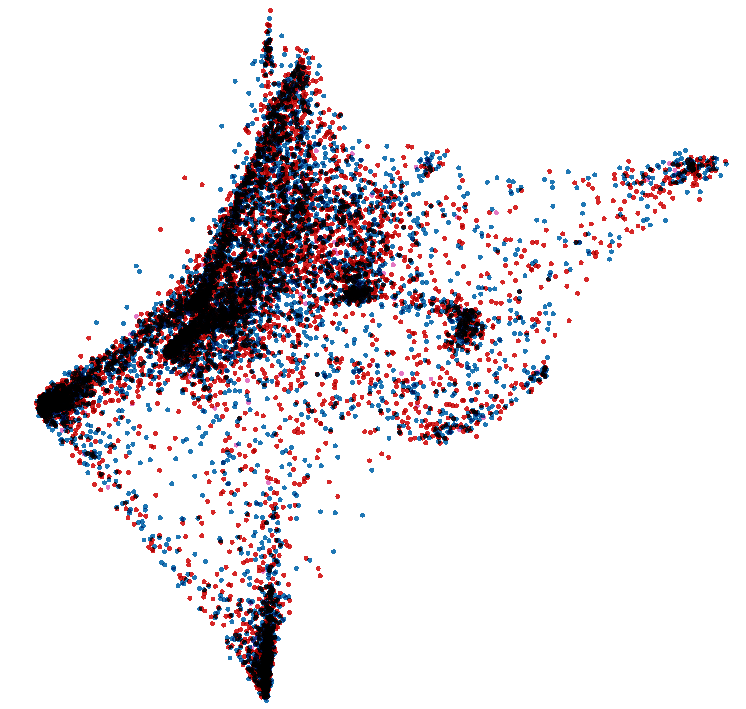}
\caption{Initial representations of users (user2vec) projected in 2D space with T-SNE. Red color denotes sarcastic users, blue non-sarcastic.}
\label{fig:initial_rep}
\end{figure}
\begin{figure}[!t]
\centering
\captionsetup{type=figure}
\includegraphics[width=\textwidth]{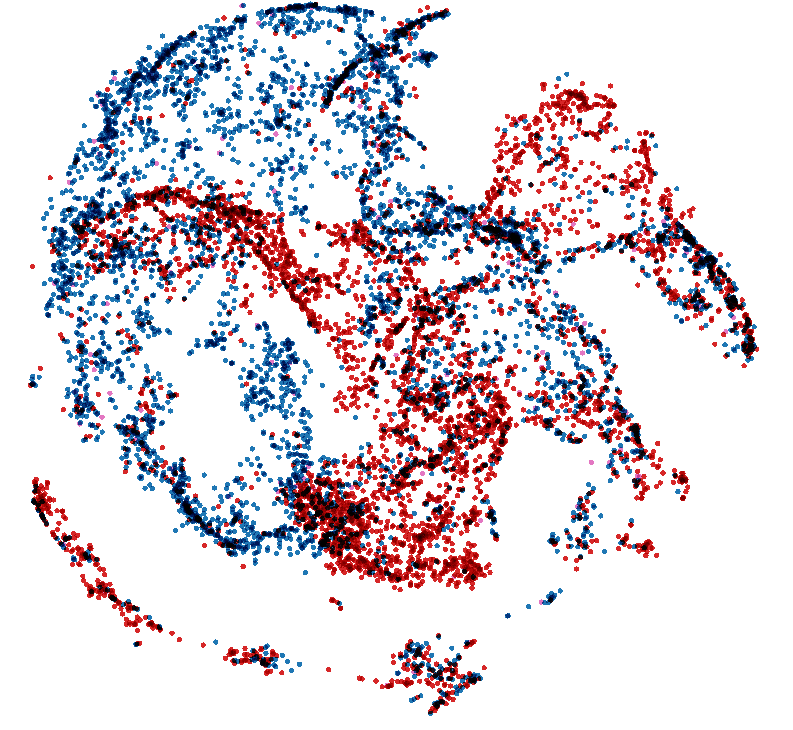}
\caption{Learned representations by our social network module (GAT) projected in 2D space with T-SNE. Red color denotes sarcastic users, blue non-sarcastic.}
\label{fig:learned_rep}
\end{figure}

\paragraph{User representation} We compare the initial user embeddings initialized by user2vec with the final representations computed from the GAT. The representations are projected in 2-dimensional space using T-SNE \cite{van2008visualizing}. In Figure \ref{fig:initial_rep} and \ref{fig:learned_rep} we visualize the initial representations with user2vec and computed representation by GAT layer respectively. While in user2vec representations sarcastic users cannot be distinguished from non-sarcastic ones, in the GAT representations we can observe communities of users sharing the same sarcastic tendency.  

\paragraph{Conversation context} For comparison, we construct three more social graphs where: 1) We remove the elicit tweets which triggered the sarcastic comment (GAT - elicit tweets), 2) We remove the oblivious tweets which interpreted the comment as serious (GAT - oblivious tweets), 3) We add the original cue tweets, revealing that the post was sarcastic (GAT + cue tweets). As expected, adding the cue tweets in the social graph leads to an almost perfect F1 score of 94.5\%. 
Removing oblivious and elicit tweets causes just a small performance drop (2-3\%). In the way the SPIRS dataset is annotated, an oblivious tweet typically triggers a cue tweet (``\textit{c mon, dude, it was just sarcasm}''). We hypothesize that even with the cue tweets removed, the model is able to learn the predictive relation between oblivious and sarcastic tweets. This is in line with the original paper (i.e. without user context), where a 3.4\% drop in prediction accuracy was observed, when the oblivious tweets were removed.

\paragraph{Attention weights} The attention mechanism of GAT is able to assign varied weights to different nodes in the neighborhood, dynamically encoding of the user by their homophily relations, which boosts the effect of authors in tweet representations \cite{flek-2020-returning}. 
We confirm this by examining users with a larger number of tweets in the dataset. When users tend to be sarcastic in most of the posts, the attention weight of their non-sarcastic tweets is smaller. In these cases, the attention weights give more importance to the surrounding user context over the conversation thread. 
Overall, the largest source of information for the model are the user nodes and the tweet that is being classified. We note the normalized attention weights are smaller for the oblivious and elicit tweet edges, and higher for the edges that connect tweets with their respective author.  
In other words, the conversational context only plays a decisive role in case of insufficient or inconsistent user-level priors.

\paragraph{Sarcasm Perception}

Cue tweets can be either authored by the same user as the sarcastic post (\textit{intended sarcasm}) or a different one (\textit{perceived sarcasm}).
We observe that in the sarcasm detection task, the error rate on perceived sarcasm is 20\% while in the self-reported sarcasm it is only around 15\%.  
We therefore test our model on distinguishing between perceived and self-reported sarcasm. Our GAT model brings an improvement of 2.2\% over the BERT baseline, with the perceived sarcasm being harder to detect (F1 56\%) than the self-reported one (F1 84\%). These results are aligned with the conclusions from \cite{oprea-magdy-2019-exploring}. In most cases, the perceived sarcasm is misclassified as self-reported, which is present more often (70\%) in the data.   
Perceived sarcasm is dependent on the readers rather than the author of the tweet, therefore we hypothesize that modeling the authors' context is less useful. It could be of benefit to model more robust recipient user profiles as well, to better predict how each individual will react. 

\begin{table}[tpb]
\small
\centering
\renewcommand{\arraystretch}{1.5}
\begin{tabular}{lccc}
\toprule
\multicolumn{4}{c}{\textbf{Sarcasm Perception}  }             \\
\midrule
\textbf{Model}                 & \textbf{P} & \textbf{R} & \textbf{F1} \\
\midrule
 BERT                          & 73.2\%     & 68.0\%     & 69.0\%        \\
                                    \textbf{User+tweet GAT}  (no cues)                & 75.0\%     & 67.7\%     & 71.2\%       \\
                                    \bottomrule
\end{tabular}
\caption{Mean overall precision (P), recall (R), and F1 score (F1) over 10 runs classifying self-reported (intended) and perceived sarcasm on the SPIRS dataset.}
\label{tab:perc_results}
\end{table}

\paragraph{Limitations} Modeling the social networks with GAT is affected by several factors. First, the low graph density, as the original dataset wasn't collected by following relationships between users, hence many users across different conversation threads are not related to each other. Second, the homophily degree is only 32\%, users with sarcastic tendency have few connections among them.


\section{Conclusions}

In this work, we explore social networks of user interactions, and contextual information to interpret sarcastic intentions in social media. We propose a graph attention-based model, which combines contextual information of users, linguistic features, and social networks. The heterogeneous social network modeling dynamically exploits relationships between users and tweets in a conversation and significantly improves the state-of-the-art results. 

\section*{Acknowledgements}
This work has been supported by the German Federal Ministry of Education and Research (BMBF) as a part of the Junior AI Scientists program under the reference 01-S20060. We would like to thank Flora Sakketou and all the anonymous reviewers for their valuable input.

\section*{Ethical Considerations}
The ability to automatically approximate personal characteristics of online users in order to improve natural language classification algorithms requires us to consider a range of ethical concerns, including: (1) privacy and user consent, (2) representativeness of the data for generalization, and (3) user vulnerability to a potential model or data misuse or misinterpretation.

Use of any user data for personalization shall be transparent, and limited to the given purpose, no individual posts shall be republished \citep{hewson2013ethics}. Researchers are advised to take account of users’ expectations \citep{williams2017towards,shilton2016we,townsend2016social} when collecting public data such as Twitter. In this case, when we expand the original dataset with more extensive user history, we utilize publicly available Twitter data in a purely observational~\citep{DBLP:journals/corr/NorvalH17}, and non-intrusive manner. All user data is kept separately on protected servers, linked to the raw text and network data only through anonymous IDs.

\citet{shah-etal-2020-predictive} identify four different sources of bias in NLP models: selection bias, label bias, model overamplification, and semantic bias. While we can't exclude any of those, the selection bias should be kept in mind in particular, when reusing the presented model, as it is unclear to which extent the augmented SPIRS dataset with user history represents a sample of the overall population on Twitter. The user selection was based solely on the available sarcasm annotations, and doesn't include any sociodemographic information. 

In addition, any user-augmented classification efforts risk invoking  stereotyping and  essentialism, as the algorithm may lean towards label people rather than posts (e.g. ``this is a sarcastic person''). Such stereotypes can cause harm even if they are accurate on average differences \citep{rudman2008social}. These can be emphasized by the semblance of objectivity created by the use of a computer algorithm \citep{koolen-van-cranenburgh-2017-stereotypes}. It is important to be mindful of these effects when interpreting the model results in an own end-application context.

\bibliography{emnlp2021}
\bibliographystyle{acl_natbib}

\newpage
\appendix
\section*{Appendix}
\section{Configurations}

We perform a stratified 90/10 train-test split. We sample $10\%$ of the training data for validation. All splits have the same class distribution and different sets of tweet authors. We use 3 GAT layers, with number of heads $K = 4$. The initial dimension is $d = 400$ and the final output dimension $d^\prime = 100$. To train our model we set learning rate to $1e-4$, and dropout 0.4 \cite{srivastava2014dropout}, and use the Adam optimization algorithm \cite{kingma2015adam} for 500 training epochs with early stopping. For the GAT layers, we compute the mean of the outputs from each attention head instead of concatenation. All experiments are run in Nvidia A100 40 GB GPUs.

\section{User Context}

To incorporate user context, we first extract all user IDs for all the tweets in the dataset. In the dataset, due to different tweet types with different users, we get in total 57K users. We fetch the tweet post timeline for each user, and we end up with a total of 104M tweets, in average 1800 posts per user. For user2vec training, we take into account only the users with a minimum of 50 posts in their timeline, and we limit the total number of posts to 1000. After filtering, the amount of tweets in the context is 10M. Every tweet is pre-processed by removing all links, user mentions are replaced with "@user", emojis and hashtags are cleared.  We train user2vec  \cite{amir-etal-2016-modelling} for 12 epochs, with learning rate 1e-4. For those users which are filtered, or we cannot extract history, we initialize them as the mean representation of his user neighbors in the social network.  We used the history tweets only for creating the user-to-user edges, and those are not present in the constructed graph, but are already encoded in the initial user representation. We experimented with various history length settings, and found almost no difference in the performance between using interactions throughout all history and interactions during the last year. Hence, we omitted older interactions to ease the computations.

\label{sec:appendix}

\end{document}